\documentclass[10pt,twocolumn,letterpaper]{article}

\usepackage{cvpr}              

\usepackage{graphicx}
\usepackage{amsmath}
\usepackage{amssymb}
\usepackage{booktabs}

\usepackage{bm}
\usepackage{multirow}
\usepackage{makecell}
\usepackage{amssymb}
\usepackage{bbding}

\usepackage[pagebackref,breaklinks,colorlinks]{hyperref}

\usepackage[capitalize]{cleveref}
\crefname{section}{Sec.}{Secs.}
\Crefname{section}{Section}{Sections}
\Crefname{table}{Table}{Tables}
\crefname{table}{Tab.}{Tabs.}

\def\cvprPaperID{5098} 
\def\confName{CVPR}
\def\confYear{2022}

\begin{document}

\title{GraftNet: Towards Domain Generalized Stereo Matching with a Broad-Spectrum and Task-Oriented Feature}
\author{Biyang Liu $^{1,2}$, Huimin Yu $^{1,2,3, 4}$, Guodong Qi $^{1,2}$
\\
  $^{1}$College of Information Science and Electronic Engineering, Zhejiang University \\ $^{2}$ZJU-League Research \& Development Center, $^{3}$State Key Lab of CAD\&CG, Zhejiang University 
 \\ $^{4}$Zhejiang Provincial Key Laboratory of Information Processing, Communication and Networking\\
{\tt\small \{biyangliu, yhm2005, guodong\_qi\}@zju.edu.cn}
}
\maketitle

\begin{abstract}
Although supervised deep stereo matching networks have made impressive achievements, the poor generalization ability caused by the domain gap prevents them from being applied to real-life scenarios. In this paper, we propose to leverage the feature of a model trained on large-scale datasets to deal with the domain shift since it has seen various styles of images. With the cosine similarity based cost volume as a bridge, the feature will be grafted to an ordinary cost aggregation module. Despite the broad-spectrum representation, such a low-level feature contains much general information which is not aimed at stereo matching. To recover more task-specific information, the grafted feature is further input into a shallow network to be transformed before calculating the cost. Extensive experiments show that the model generalization ability can be improved significantly with this broad-spectrum and task-oriented feature. Specifically, based on two well-known architectures PSMNet and GANet, our methods are superior to other robust algorithms when transferring from SceneFlow to KITTI 2015, KITTI 2012, and Middlebury. Code is available at \emph{https://github.com/SpadeLiu/Graft-PSMNet}.
\end{abstract}

\section{Introduction}

As a low-cost means to acquire depth, stereo matching has been studied as a fundamental problem in the vision society for decades. Given a rectified image pair, the objective is to search for the corresponding points and calculate their disparities. Stereo matching algorithms generally involve four steps \cite{scharstein2002taxonomy}: matching cost computation, cost aggregation, disparity optimization, and disparity refinement.

Although Convolutional Neural Network (CNN) based supervised stereo matching methods have achieved admirable performances, huge amounts of annotated data are required to train the models, which is cumbersome for real-life applications. Synthetic data \cite{mayer2016large} is sufficient while the domain gap between the source and the target images prevents the models from generalizing well. There are three solutions to this issue: the unsupervised image reconstruction loss \cite{tonioni2019real, tonioni2019learning}, domain adaptation techniques \cite{liu2020stereogan, song2021adastereo}, and domain generalized approaches \cite{zhang2020domain, cai2020matching}. In this paper, we focus on the third situation which is more challenging since the target images are not available during training.

\begin{figure}[t]
\centering
\includegraphics[width=0.9\linewidth]{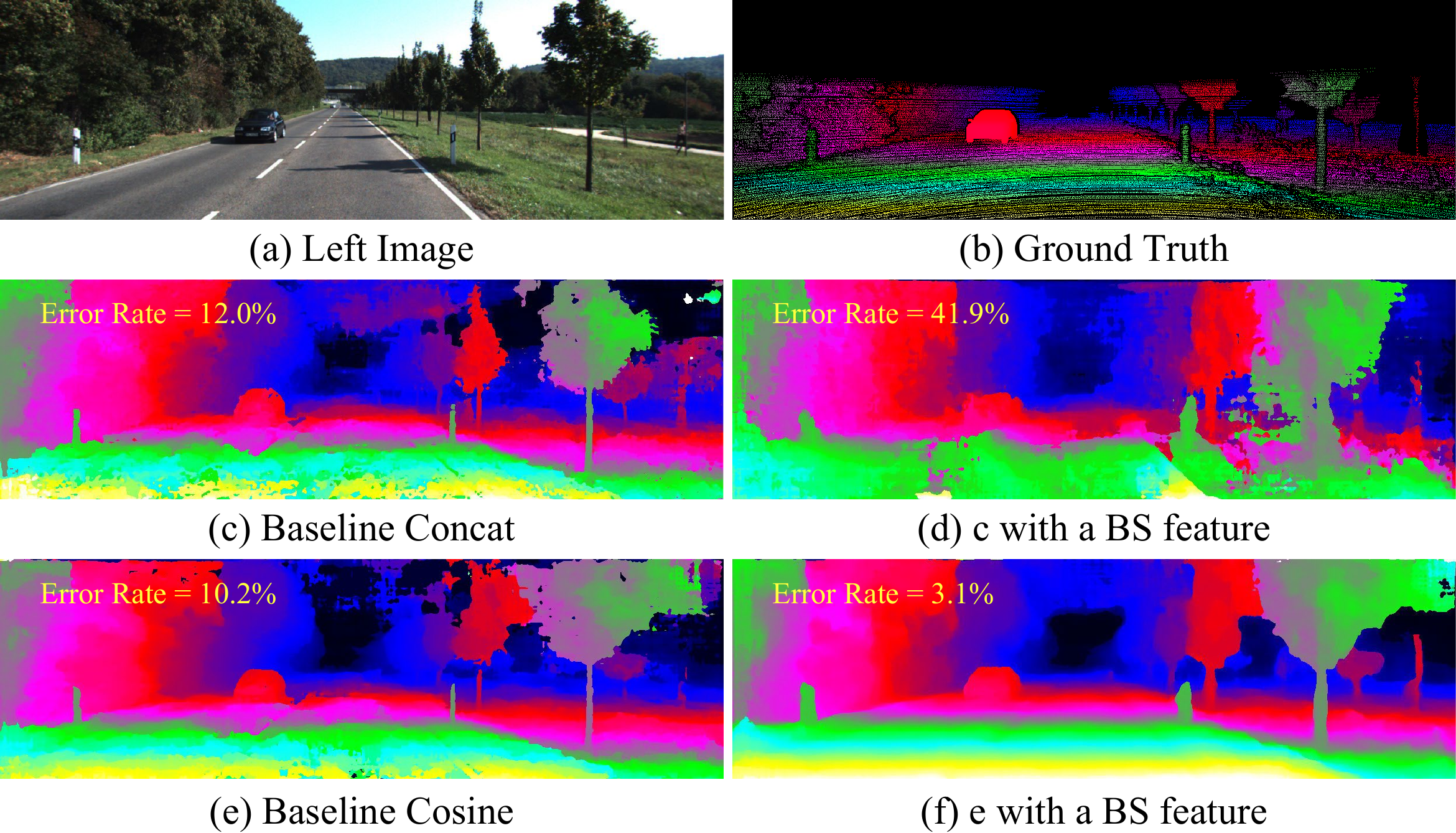}
\caption{The toy experiment to validate the grafting operation. Two models with the cost volume formed by feature concatenation (Subfigure \textbf{c}) and cosine similarity (Subfigure \textbf{e}) are trained on SceneFlow, then their feature extraction modules are replaced with a \textbf{B}road-\textbf{S}pectrum feature (Subfigure \textbf{d} and \textbf{f}). For the four models, the 3-pixel error rates on a KITTI sample are labeled.}
\label{introduction}
\end{figure}

In domain generalized stereo matching, feature representation plays a crucial role \cite{zhang2020domain} since the feature extraction module directly confronts images from different domains. Then a question is raised: can the goal be achieved by replacing the feature extraction module of an ordinary stereo matching network (\emph{ordinary} means it is trained with synthetic data) with a broad-spectrum feature (\ie the feature of a model trained on large-scale datasets)? Since this feature has seen various styles of images and learns to generalize well. In traditional algorithms, various feature descriptors \cite{hirschmuller2007evaluation} and cost aggregation methods \cite{yoon2006adaptive, hirschmuller2005accurate} could be combined with each other to use. However in deep frameworks, the parameterized modules are entangled through end-to-end training, is this grafting operation (\ie combining two trained modules without finetuning) practical?

To answer this question, we first conduct a toy experiment. With PSMNet \cite{chang2018pyramid} as the basic architecture, we train a model on a synthetic dataset SceneFlow \cite{mayer2016large}, then its feature extraction module is replaced with the feature of VGG \cite{simonyan2014very} trained on ImageNet \cite{deng2009imagenet}. Finally, the cross-domain performances are evaluated on KITTI 2015 \cite{menze2015object}. As illustrated in Subfigure (c) and (d) of Figure \ref{introduction}, simply grafting a broad-spectrum feature to an ordinary cost aggregation module leads to a collapse of the disparity result. We analyze this is caused by the feature concatenation based cost volume, which forces the cost aggregation module to learn to measure the similarity based on the feature. When the feature is replaced, the learned metric will not be effective.

In order to disentangle the feature extraction module and the cost aggregation module, it is necessary to construct a generalized cost space \cite{cai2020matching}. On one hand, the cost volume should contain pure similarity information. In this way, the prior knowledge about the similarity metric is injected, preventing the cost aggregation module from overfitting to the used feature. Besides, the semantical information \cite{guo2019group} which may interfere with cost aggregation due to the varied semantic classes of different domains is discarded. On the other hand, integrating the normalization of the cost value is beneficial for the generalization ability \cite{song2021adastereo}. To this end, we utilize the elegant \emph{cosine similarity} to construct the cost volume. In addition to satisfying the above demands, cosine similarity projects features with arbitrary channels to a scalar, making the cost accessible for various features.

Owing to the generalized cost space, when the feature extraction module trained with synthetic data is replaced with a broad-spectrum feature, the cross-domain performance is improved significantly, as shown in Subfigure (e) and (f) of Figure \ref{introduction}. This also experimentally validates that a broad-spectrum feature can be employed to handle the domain shift. However, grafting such a low-level feature of the classification model is still suboptimal since it contains much general information that serves various tasks. It is necessary to adapt the grafted feature to our stereo matching task. Inspired by the researches in multi-task learning \cite{li2020knowledge} and transfer learning \cite{ramirez2019learning}, we build a shallow network and force it to recover more task-specific information from the grafted feature. Although this training process is conducted on the source domain, the feature adaptor is robust since its input, the broad-spectrum feature has weakened the influence of the image style. Besides, the small amount of the parameters will reduce the risk of overfitting \cite{wang2020generalizing}.

In summary, there are two fundamental steps in our domain generalized stereo matching network GraftNet. Firstly, grafting a broad-spectrum feature (\ie the feature of a model trained on large-scale datasets) to the cost aggregation module of an ordinary stereo matching network. Secondly, transforming the feature with a shallow network to recover the task-specific information. In practice, we find retraining the cost aggregation module with this transformed feature can further improve the performance. It is worth noting that our method can be built upon arbitrary stereo matching networks, the only modification is to construct the cost volume with cosine similarity. Without bells and whistles, our models based on PSMNet \cite{chang2018pyramid} and GANet \cite{zhang2019ga} are superior to other robust and domain generalized algorithms when transferring from a synthetic dataset SceneFlow \cite{mayer2016large} to some realistic datasets such as KITTI 2015 \cite{menze2015object}, KITTI 2012 \cite{geiger2012we}, and Middlebury \cite{scharstein2014high}.

\section{Related Work}

\subsection{Deep Stereo Matching Networks}

MC-CNN \cite{zbontar2015computing} firstly introduced CNN to stereo matching, where a siamese network was built to compute the matching cost of two patches. The subsequent studies involved multi-scale feature representation \cite{chen2015deep} and the acceleration of the similarity calculation \cite{luo2016efficient}. Although a deep embedding is powerful, these works are limited by the traditional cost aggregation and disparity refinement steps.

DispNetC \cite{mayer2016large} was the first end-to-end stereo matching network, where the disparity was regressed from the correlation maps through 2D convolutions. This pipeline has been widely adopted since then. SegStereo \cite{yang2018segstereo} and EdgeStereo \cite{song2020edgestereo} designed multi-task frameworks to exploit the semantic clues and the edge information. AANet \cite{xu2020aanet} integrated deformable convolution to adaptively aggregate the cost. Despite the low complexities, the performances of 2D-CNN based stereo matching networks are not superior.

Another common fashion is to build the cost volume by concatenating the left and the right features and aggregate the cost with 3D convolutions, which is initially proposed in GCNet \cite{kendall2017end}. PSMNet \cite{chang2018pyramid} further introduced the spatial pyramid pooling module to deal with textureless regions. For more effective cost aggregation, image content guided layers were designed in GANet \cite{zhang2019ga}. Currently, LEAStereo \cite{cheng2020hierarchical}, an architecture searched by a deep network, ranks top on the KITTI benchmarks. While 3D-CNN based stereo matching networks perform well on several datasets, the poor generalization abilities hinder their applications in real-life scenes. In this work, we show how to alleviate this problem with a domain-invariant and task-oriented feature.

\begin{figure*}
  \centering
\includegraphics[width=0.8\linewidth]{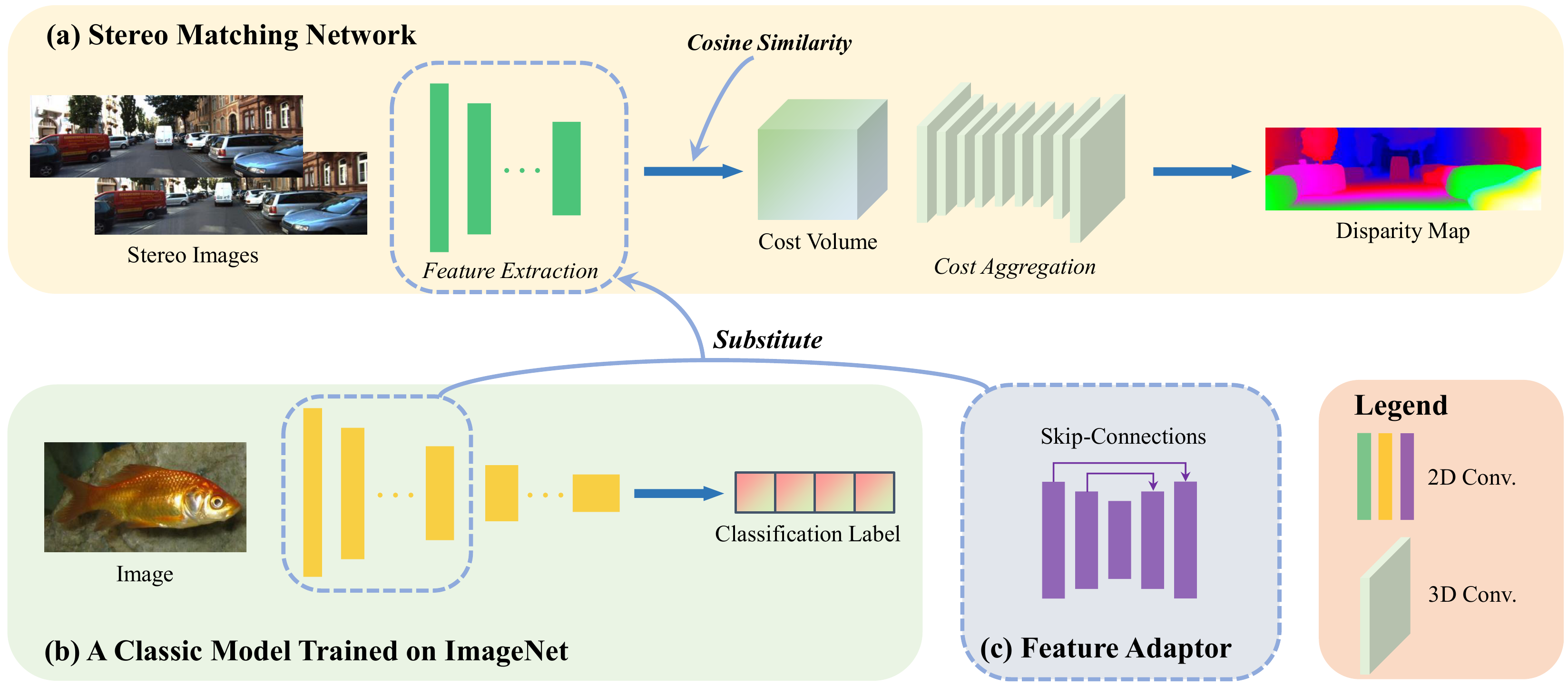}
\caption{The overall architecture of GraftNet, consisting of a broad-spectrum feature, a feature adaptor, and a cost aggregation module. To build GraftNet, we first graft the feature from a classic model trained on the large-scale dataset ImageNet to the cost aggregation module of an ordinary stereo matching network. Then the feature is input to a shallow U-shape network to be transformed to recover more task-related information. Note the cost volume is formed by cosine similarity rather than feature concatenation to obtain a generalized cost space.}
\label{architecture}
\end{figure*}

\subsection{Domain Generalized Stereo Matching}

In domain generalized stereo matching, the model is agnostic to the image style of the target domain, thus it is more challenging than domain adaptation \cite{tonioni2019real, tonioni2019learning, liu2020stereogan, song2021adastereo}. To achieve the goal, DSMNet \cite{zhang2020domain} proposed domain normalization and a structure-preserving graph-based filter. CFNet \cite{shen2021cfnet} adaptively adjusted the search space to deal with the unbalanced disparity distribution across different domains. STTR \cite{li2021revisiting} and RAFT-Stereo \cite{lipson2021raft} introduced novel architectures which showed strong robustness. Our method is most similar to MS-Net \cite{cai2020matching}, where traditional descriptors are utilized to construct the cost. However, in our work, the generalized matching space is realized with a deep feature which is more discriminative.

In addition, there are researchers devoted to tackling the problem from other perspectives. Poggi \etal \cite{poggi2019guided} modulated the cost distribution with the sparse depth measurements obtained from some devices. Watson \etal \cite{watson2020learning} proposed an approach to generate labeled data from single images and showed that models trained on their MfS transferred better than those trained on SceneFlow \cite{mayer2016large}.

\subsection{Broad-Spectrum Features}

The features of the classic architectures (\eg ResNet \cite{he2016deep}, VGG \cite{simonyan2014very}) trained on ImageNet \cite{deng2009imagenet} have been widely utilized to initialize the model parameters in several tasks \cite{chen2017deeplab, godard2019digging}. These pretrained classic models can be easily loaded from the libraries, \eg PyTorch \cite{paszke2017automatic}. In our work, the feature is leveraged to obtain a robust representation since ImageNet covers various domains. To preserve the property of the feature, we keep the parameters fixed and build a network to transform the feature \cite{li2020knowledge, ramirez2019learning}.


\section{Method}

In this section, we will describe how to build our domain generalized stereo matching model GraftNet, whose key component is a broad-spectrum and task-oriented feature. The overall architecture is illustrated in Figure \ref{architecture}. Specifically, we first train a stereo matching network with the cost volume formed by cosine similarity (Section \ref{section3.1}). Then we graft the feature from a classic model trained on ImageNet to the cost aggregation module of this stereo matching network and further transform it with a shallow network to recover the task-related information (Section \ref{section3.2}). Finally, we empirically retrain the cost aggregation module with the transformed feature (Section \ref{section3.3}).

\subsection{Stereo Matching Network}
\label{section3.1}

In a typical deep stereo matching network \cite{kendall2017end, chang2018pyramid, zhang2019ga, cheng2020hierarchical}, the left and the right images are first passed through the feature extraction module, then a cost volume is constructed by concatenating the left and the right features at different displacements. After that, the cost is aggregated with several 3D convolutions, followed by \emph{softmax} and \emph{weighted average} to calculate the final disparity. 

As demonstrated in \cite{zhang2020domain}, feature representation plays a crucial role in the generalization ability of the model. To this end, we intend to achieve the domain generalized stereo matching with a broad-spectrum feature. At the same time, the other parameterized part of a stereo matching network, the cost aggregation module, can only be trained with the synthetic data. Therefore, it is necessary to construct a generalized cost space \cite{cai2020matching} to disentangle the feature extraction module and the cost aggregation module.

In our model, the elegant \emph{cosine similarity} is utilized to build the cost volume. Compared with feature concatenation, it has three advantages: 1) The semantical information \cite{guo2019group} which is susceptible to the domain shift is eliminated, resulting in a cost volume with pure similarity information. 2) The normalization ensures the numerical stability of the cost values, which is beneficial for the cross-domain evaluation performance \cite{zhang2020domain, song2021adastereo}. 3) Features with arbitrary channels could be taken as the input since all of them will be projected to a scalar. Formally, the cosine similarity cost volume is expressed as: 
\begin{equation}
\mathbf{CV}_{cos}(:, d, x, y) = \frac{\langle F^l(:, x, y), F^r(:, x-d, y) \rangle}{||F^l(:, x, y)||_2 \cdot ||F^r(:, x-d, y)||_2}
\end{equation}
where $d$ is the disparity index, and $(x, y)$ denotes the pixel coordinate. $F^l$ and $F^r$ are the left and the right features, both with $C$ channels. The calculated cost is a 4D tensor with only one channel, thus the input channel of the first 3D convolutional layer of the cost aggregation module should be modified to 1.

Without other adjustments, this basic stereo matching network is trained on the source domain with the cross entropy loss \cite{tulyakov2018practical} and the smooth $L1$ loss \cite{chang2018pyramid} to supervise the disparity probability distributions and the final disparity values respectively:
\begin{equation}
L_{ce}(\hat{P}(d), P(d)) = \frac{1}{N}\sum_{i=1}^{N} \sum_{d=0}^{d_{max}} -\hat{P_i}(d)\cdot \log P_i(d)
\end{equation}
\begin{equation}
L_{sm}(\hat{D}, D) = \frac{1}{N}\sum_{i=1}^{N} smooth_{L_1}(\hat{D_i}, D_i)
\end{equation}
where $\hat{P}(d)$ is the predicted distribution from \emph{softmax} and $P(d)$ is the ground truth distribution, a normalized Laplacian distribution centered at the disparity ground truth $D$. $\hat{D}$ is the predicted disparity calculated by \emph{weighted average}. $N$ denotes the number of the pixels in an image.

Normally, there are multiple disparity results output from the cost aggregation module in the training phase \cite{chang2018pyramid, zhang2019ga, guo2019group}. In our model, each result is supervised with the above two loss functions, then the total loss is:
\begin{equation}
L = \sum_{m=1}^{M} \lambda_m(L_{ce}+\mu L_{sm})
\end{equation} 
where $M$ is the number of the disparity outputs. As for the balance weights, $\lambda_m$ is set as same as in the adopted basic architecture, and $\mu$ is set to 0.1 heuristically. 

After training, the feature extraction module is discarded since it is susceptible to the domain shift, while the cost aggregation module is reserved for grafting other features. Owing to the generalized cost space, the cost aggregation module is less affected by the domain gap.

\subsection{Broad-Spectrum and Task-Oriented Feature}
\label{section3.2}

In this work, we employ the feature of a model trained on large-scale datasets to resist the domain shift since it has seen various styles of images and has learned to generalize well. In the meanwhile, such a feature is easy to acquire, \eg the classic models \cite{simonyan2014very, he2016deep} trained on ImageNet \cite{deng2009imagenet} can be directly loaded from the PyTorch library \cite{paszke2017automatic}. Rather than utilizing the pretrained parameters to initialize the model backbones \cite{chen2017deeplab, godard2019digging}, we keep the module fixed to preserve the inherent property of the feature.

Specifically, the broad-spectrum feature will be grafted to the trained ordinary cost aggregation module in Section \ref{section3.1}. To keep consistency, we adopt the feature that has the same resolution as the one used in the original stereo matching network. For example, if the basic architecture is PSMNet \cite{chang2018pyramid} and the grafted feature is from VGG \cite{simonyan2014very}, then the feature before the third pooling layer that has the quarter resolution of the image will be employed.

Although the influence of the domain shift has been weakened by a broad-spectrum feature, a simple grafting operation is ill-considered. The reason is that the feature is relatively low-level, containing much general information that serves various downstream tasks. It is necessary to extract more information specific to our stereo matching task.

To this end, we build a feature adaptor before calculating the cost, \ie the shallow U-shape network \cite{ronneberger2015u} illustrated in Figure \ref{architecture} (c). The feature adaptor is trained as a part of the stereo matching network, in which process the parameters of the broad-spectrum feature and the cost aggregation module are fixed and only serve as the intermediums to propagate the gradients. Although this training process is conducted on the source domain, the feature adaptor is effective on the target domain for two reasons: 1) Its input is a broad-spectrum representation which will weaken the influence of the image style. 2) The small amount of the parameters will reduce the risk of overfitting \cite{wang2020generalizing}.

\subsection{GraftNet}
\label{section3.3}

With the broad-spectrum and task-oriented feature output from the feature adaptor, we find retraining the cost aggregation module can further improve the performance. In this step, our method is similar to \cite{cai2020matching}, \ie constructing a generalized matching space and training a cost aggregation module with the synthetic data. However, experimental results in Section \ref{section4.6} show that an appropriate deep feature is more representative than traditional descriptors \cite{cai2020matching}. 

From the perspective of the model architecture, GraftNet consists of three components: a broad-spectrum feature, a feature adaptor, and a cost aggregation module. Although we are inspired by the toy grafting experiment in Figure \ref{introduction}, can the feature adaptor and the cost aggregation module be trained together? In practice, we find jointly training is not as effective as separately training (Please refer to the supplementary material). We conjecture that when the two modules are optimized individually, a trained module can provide a beneficial initialization for the other one.

\begin{table*}\small
\centering
\begin{tabular}{lcccccccccc}
\toprule
\multirow{2}{*}{\textbf{~Model}} & \multirow{2}{*}{\textbf{Step}} & \multicolumn{2}{c}{KITTI 2015} & \multicolumn{2}{c}{KITTI 2012}  & \multicolumn{2}{c}{Middlebury} & \multicolumn{2}{c}{ETH3D}  \\
\cmidrule(l){3-4}
\cmidrule(l){5-6}
\cmidrule(l){7-8}
\cmidrule(l){9-10}
& & EPE (px) & $>$3px & EPE (px) & $>$3px & EPE (px) & $>$2px& EPE (px) & $>$1px	 \\
\midrule
\multirow{5}{*}{PSMNet} & \makecell[l]{Baseline}				& 3.24 & 19.5\% & 2.59 & 18.6\% & 6.69 & 22.6\% & 2.20 & 12.1\% \\
				   & \makecell[l]{Cosine Similarity CV} 	& 2.98 & 15.4\% & 2.30 & 14.3\% & 6.83 & 22.3\% & 1.17 & \textbf{10.6\%} \\
				   & \makecell[l]{Graft VGG's Feature}	 	& 1.86 & 6.39\% & 1.28 & 5.90\% & 5.67 & 18.9\% & 1.81 & 11.9\% \\
				   & \makecell[l]{+ Feature Adaptor} 		& 1.47 & 5.60\% & 1.16 & 5.20\% & 2.96 & 12.0\% & 1.66 & 12.6\% \\
				   & \makecell[l]{Retrain CA Module} 		& \textbf{1.32} & \textbf{5.34\%} & \textbf{1.09} & \textbf{4.97\%} & \textbf{2.34} & \textbf{10.9\%} & \textbf{1.16} & 10.7\% \\
\midrule
\multirow{5}{*}{GANet}  & \makecell[l]{Baseline}				& 2.76 & 17.1\% & 2.35 & 12.8\% & 7.33 & 20.7\% & 0.46 & 7.80\% \\
				  & \makecell[l]{Cosine Similarity CV} 		& 1.80 & 8.78\% & 1.70 & 8.57\% & 6.09 & 21.3\% & 0.46 & 7.08\% \\
				  & \makecell[l]{Graft VGG's Feature} 		& 1.91 & 7.12\% & 1.91 & 8.37\% & 7.75 & 24.3\% & 0.86 & 13.2\% \\
				  & \makecell[l]{+ Feature Adaptor} 		& 1.31 & 5.55\% & 1.19 & 5.11\% & 1.96 & 10.9\% & 0.45 & 6.67\% \\ 
				  & \makecell[l]{Retrain CA Module} 		& \textbf{1.30} & \textbf{5.35\%} & \textbf{1.07} & \textbf{4.60\%} & \textbf{1.87} & \textbf{8.89\%} & \textbf{0.43} & \textbf{6.17\%} \\ 
\bottomrule
\end{tabular}
\caption{Quantitative results of the ablation experiment. PSMNet and GANet-11 are the used two basic architectures. Models are trained on SceneFlow and evaluated on four realistic datasets. CV represents cost volume and CA denotes the cost aggregation module.}
\label{ablation}
\end{table*}

Since grafting is the first and fundamental step in the whole pipeline, our domain generalized stereo matching network is termed GraftNet. Moreover, we wish the grafting operation could provide a novel viewpoint: Can parts of two trained CNNs be integrated without finetuning to obtain a new model? This question is worth exploring, especially for scenarios where training data is not available.


\section{Experiment}

\subsection{Datasets \& Evaluation Metrics}

\textbf{Source Domain.} In the experiment, all of the stereo matching networks are trained on \textbf{SceneFlow} \cite{mayer2016large}, a synthetic dataset containing 35454 training pairs and 4370 testing pairs, both with dense disparity ground truth. Since only the generalization ability is concerned in the domain generalized problem, the test set will not be used.

\textbf{Target Domain.} The models trained on SceneFlow are evaluated on the following realistic datasets:
\begin{itemize}
\item \textbf{KITTI} datasets consist of KITTI 2015 \cite{menze2015object} and KITTI 2012 \cite{geiger2012we}, whose ground truth disparity maps are sparse. On KITTI 2015, there are 200 training pairs and 200 testing pairs. On KITTI 2012, there are 194 training pairs and 195 testing pairs.
\item \textbf{Middlebury 2014} \cite{scharstein2014high} provides 15 training pairs and 15 testing pairs, where some samples are under inconsistent illumination or color conditions. All of the images are available in three different resolutions, we select the half-resolution ones.
\item \textbf{ETH3D} \cite{schoeps2017cvpr} is a gray-scale dataset with 27 training pairs and 20 testing pairs.
\end{itemize}

For all the realistic datasets, we use their training sets to evaluate the cross-domain performance. The utilized metrics are \textbf{EPE} (End Point Error, the mean average error) and \textbf{$\bm{\tau}$-pixel error rate} (percentage of the points with absolute error larger than $\tau$ pixel).

\subsection{Implementation Details}

The framework is implemented on PyTorch \cite{paszke2017automatic}, with Adam ($\beta_1=0.9$, $\beta_2=0.999$) as the optimizer. For the basic stereo matching architecture, we train it for 8 epochs with a learning rate of 0.001. Then a broad-spectrum feature is grafted to the cost aggregation module, in which process no training is involved. After grafting, the feature adaptor is trained for 1 epoch with a learning rate of 0.001. Finally, the cost aggregation module is retrained for 10 epochs with the learning rate set as 0.001 for the first 5 epochs and 0.0001 for the remaining epochs.

For all experiments, PSMNet \cite{chang2018pyramid} is adopted as the basic architecture. In the ablation study (Section \ref{section4.3}) and the comparison experiment with other robust algorithms (Section \ref{section4.6}), GANet-11 \cite{zhang2019ga} is additionally utilized to demonstrate the effectiveness and versatility of our method. The grafted feature is from VGG16 \cite{simonyan2014very} which is trained on ImageNet \cite{deng2009imagenet}, and in Section \ref{section4.4} more features are explored.

\begin{figure*}
  \centering
\includegraphics[width=0.9\linewidth]{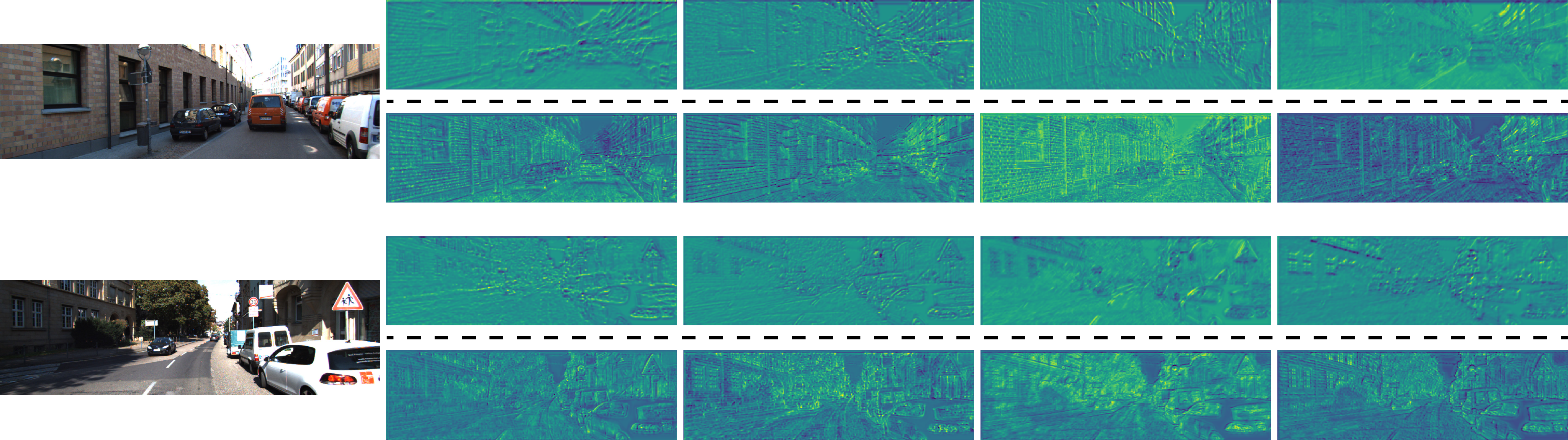}
\caption{Illustration of the adapted features. For each image, the top row shows four channels of the features before adaption (\ie VGG's feature) and the bottom row shows four channels of the features after adaption. Images are from KITTI 2015.}
\label{feature}
\end{figure*}

\begin{table}\small
\centering
\begin{tabular}{ccccc}
\toprule
Cost & Normal. & P. Simi. & EPE (px) & $>$3px \\
\midrule
\makecell[l]{Concat} & \makecell[c]{\XSolidBrush} & \makecell[c]{\XSolidBrush} & 3.24 & 19.5\% \\
\makecell[l]{N\_Concat} & \makecell[c]{\Checkmark} & \makecell[c]{\XSolidBrush} & 3.14 & 18.3\% \\
\makecell[l]{L2 Distance} & \makecell[c]{\XSolidBrush} & \makecell[c]{\Checkmark}  & \textbf{2.86} & 16.2\% \\
\makecell[l]{Cosine Simi.}  & \makecell[c]{\Checkmark} & \makecell[c]{\Checkmark} & 2.98 & \textbf{15.4\%} \\
\bottomrule
\end{tabular}
\caption{Effects of the manner of building the cost volume. \textbf{Normal.}: whether the features are normalized before building the cost. \textbf{P. Simi.}: whether the cost contains pure similarity information. \textbf{N\_Concat} means the cost is formed by concatenating the normalized features. Results are evaluated on KITTI 2015.}
\label{cost}
\end{table}

\subsection{Ablation Study}
\label{section4.3}

In this section, we study the effects of the components in GraftNet, the evaluation results are listed in Table \ref{ablation}. First of all, although feature concatenation is commonly used to build the cost in supervised frameworks, cosine similarity is more suitable for domain generalized stereo matching. There are two reasons: 1) the normalization keeps the values stable, 2) the semantical information which is susceptible to the domain shift is discarded.  

For the purpose of deeply investigating the influences of these two aspects, we compare several manners of building the cost in Table \ref{cost}. Results show that a cost volume containing pure similarity information (\eg calculated by cosine similarity or $L2$ distance) is better when generalization ability is considered. We also emphasize that cosine similarity allows us to graft features from other models. If the cost volume is constructed by feature concatenation, the disparity result of the assembled model will be collapsed.

From the \emph{2nd} and the \emph{3rd} rows of the two subtables in Table \ref{ablation}, grafting the feature of VGG trained on ImageNet is beneficial for KITTI datasets. This indicates the overfitting problem of current stereo matching networks does exist, and a broad-spectrum feature can improve the generalization ability. However, this feature does not work for Middlebury (when equipped with GANet) and ETH3D. We analyze although a domain-generalized representation is obtained, it does not fit the task, the low-level feature from the classification model contains seldom detailed information. Coincidentally, on KITTI, the ground truth is sparse especially at the disparity discontinuities, thus a feature with more global context information is effective as well.

To restore the task-specific information, we build a shallow network to transform the broad-spectrum feature. From the \emph{4th} and the \emph{9th} rows of Table \ref{ablation}, the evaluation performances on Middlebury and ETH3D are improved significantly with the feature adaptor. This result suggests that both the \emph{broad-spectrum} property and the \emph{task-oriented} property of the feature is important. In Figure \ref{feature}, we further exhibit the features before and after adaption. As it can be seen, rich texture information which is essential for stereo matching is recovered with the adaptor.

\begin{table}\small
\centering
\begin{tabular}{cccc}
\toprule
Architecture & EPE (px) & $>$3px \\
\midrule
\makecell[l]{\XSolidBrush} & 1.86 & 6.39\% \\
\makecell[l]{Linear} & 1.50 & 6.06\%  \\
\makecell[l]{Non-Linear} & \textbf{1.46} & 6.20\% \\
\makecell[l]{U-Net}  & 1.47 & \textbf{5.60\%} \\
\bottomrule
\end{tabular}
\caption{Effects of the architecture of the feature adaptor. The first row means no adaptor is utilized. The linear adaptor is a single convolutional layer and the non-linear adaptor consists of two convolutional layers and an activation layer. U-Net is the architecture shown in Figure \ref{architecture} (c). Results are evaluated on KITTI 2015.}
\label{adaptor}
\end{table}

In Table \ref{adaptor}, we compare different architectures of the feature adaptor. Although a linear layer is enough in \cite{li2020knowledge}, in our work a more complex network is needed since not only the task gap but also the feature level should be considered. In the meanwhile, the parameter number of the adaptor can not be too large to prevent overfitting to the source data. Therefore, a shallow U-shape network \cite{ronneberger2015u} is adopted.

\begin{figure*}
  \centering
\includegraphics[width=0.9\linewidth]{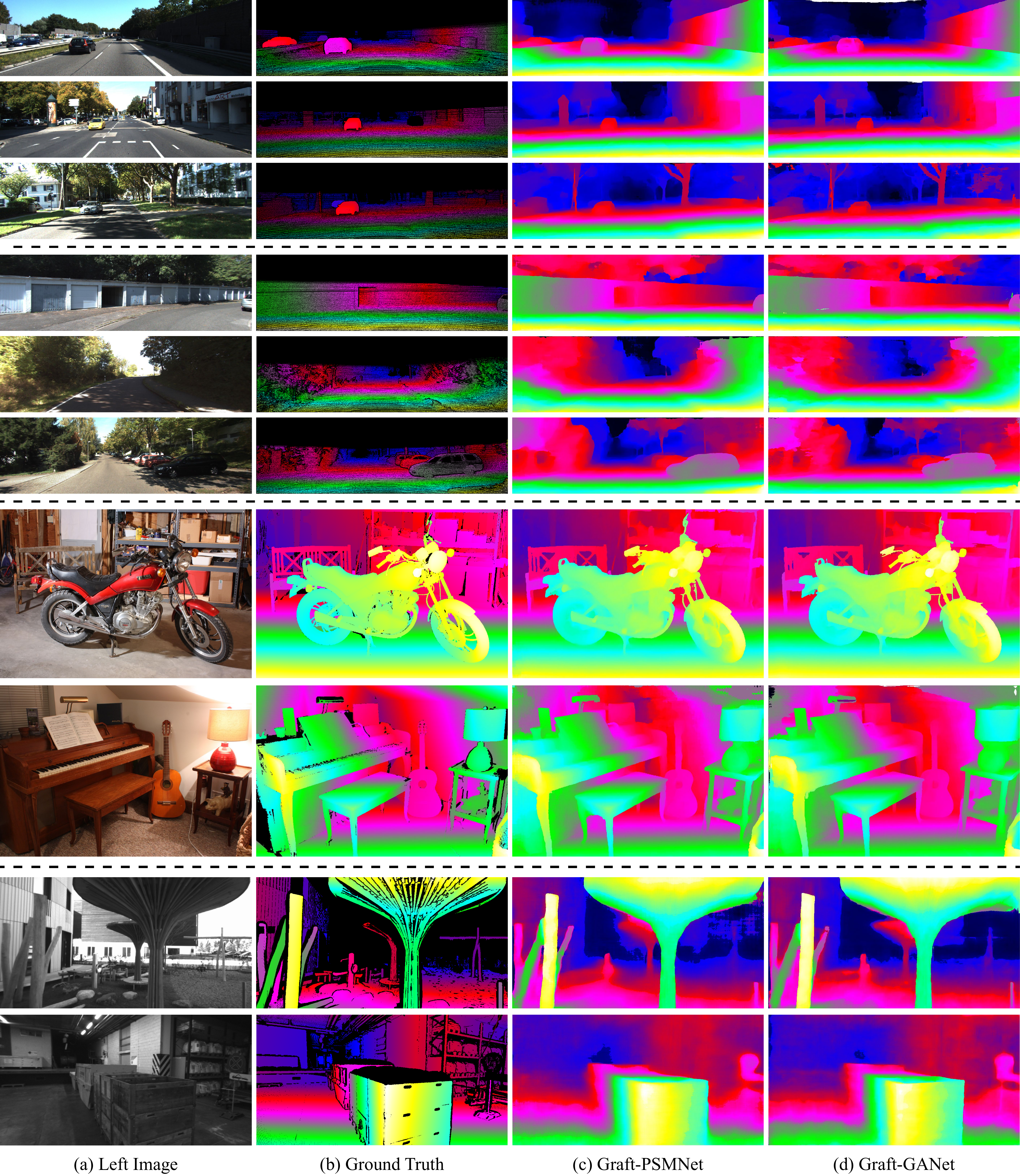}
\caption{Qualitative results of Graft-PSMNet and Graft-GANet when transferring from SceneFlow to KITTI 2015, KITTI 2012, Middlebury, and ETH3D (from top to bottom). Best viewed in color.}
\label{result}
\end{figure*}

Finally, in Table \ref{ablation}, retraining the cost aggregation module with the adapted feature can further improve the evaluation performance. The reason might be that compared with the original feature trained on the source domain, the broad-spectrum and task-oriented feature provides a more robust cost volume, guiding the optimization of the cost aggregation module towards the goal of domain generalized stereo matching. Some qualitative results of our final model on the four realistic datasets are displayed in Figure \ref{result}. 

\subsection{Grafting Various Features}
\label{section4.4}

In this section, we attempt to graft various features to further investigate the impact of the feature. Six features are adopted: VGG16 \cite{simonyan2014very}, ResNet18 \cite{he2016deep}, and ResNet50 \cite{he2016deep} trained for Classification (C) on ImageNet, ResNet18 trained for Monocular Depth Estimation (MDE) \cite{godard2019digging} on KITTI, ResNet50 trained by Dense Contrastive Learning (DCL) \cite{wang2021dense} on ImageNet, stacked ResBlocks trained for Optical Flow Estimation (OFE) \cite{teed2020raft} on KITTI.

The qualitative results evaluated on KITTI 2015 are presented in Table \ref{graft}. Comparing the \emph{2nd} row and the \emph{3rd} row, a broad-spectrum feature performs close to a domain-specific feature, meaning that the domain shift can be handled with the feature of a model trained on large-scale datasets. From the \emph{3rd} and the \emph{5th} rows, although MDE and DCL are dense prediction tasks, their features cannot satisfy the needs of stereo matching. The feature trained for a closer task OFE might be more helpful, while there is still a performance gap between it and the feature used in our model (\textbf{6.22\%} vs \textbf{5.60\%}). These conclusions once again stress the importance of the task-oriented property of the feature.

In addition, considering image classification models and stereo matching models are usually trained with different input resolutions, and the input resolution is vital for the pixelwise task stereo matching, we study the effect of the input resolution when training the broad-spectrum feature. Please refer to the supplementary material for more results.
 

\begin{table}\small
\centering
\begin{tabular}{ccccc}
\toprule
Feature & Task & Dataset & EPE (px) & $>$3px \\
\midrule
\makecell[l]{VGG16} & C & ImageNet & 1.86 & 6.39\% \\
\makecell[l]{ResNet18} & C & ImageNet &1.90 & 6.62\%  \\
\makecell[l]{ResNet18} & MDE & KITTI & \underline{1.73} & 6.54\% \\
\makecell[l]{ResNet50} & C & ImageNet & 2.06 &\textbf{6.19\%} \\
\makecell[l]{ResNet50} & DCL & ImageNet & 2.20 & 9.17\% \\
\makecell[l]{ResBlocks} & OFE & KITTI & \textbf{1.59} & \underline{6.22\%} \\
\bottomrule
\end{tabular}
\caption{Experimental results of grafting features from various models. \textbf{C}: classification, \textbf{MDE}: monocular depth estimation, \textbf{DCL}: dense contrastive learning, \textbf{OFE}: optical flow estimation. Results are evaluated on KITTI 2015, the best is shown in \textbf{bold} and the second is \underline{underlined}.}
\label{graft}
\end{table}

\subsection{Comparison with Robust Algorithms}
\label{section4.6}

In this section, we compare our models with other robust and domain generalized methods. As reported in \cite{watson2020learning, li2021revisiting}, augmenting images with a random color and brightness transform can improve the model generalization ability. Therefore, for a fair comparison, the algorithms are separated into two categories according to whether data augmentation strategies including \emph{color jitter} are involved.

As shown in Table \ref{comparison}, among the methods that do not utilize the random color transform strategy, our Graft-PSMNet and Graft-GANet are superior, especially on KITTI and Middlebury. When integrating more data augmentation approaches, the model performance can be further boosted. On ETH3D, our models are not the best, we analyze the reason is that ImageNet contains few gray-scale images, making the grafted feature difficult to express well on ETH3D. This inspires us that more styles of images should be collected for an absolutely domain-invariant representation.

\begin{table}\small
\centering
\begin{tabular}{lcccc}
\toprule
\multirow{2}{*}{~~~~Model} & KT-15 & KT-12 & MB & ET \\
					& $>$3px & $>$3px & $>$2px & $>$1px \\
\midrule
\makecell[l]{GwcNet \cite{guo2019group}} 				& 22.7\% & 20.2\% & 37.9\% & 54.2\% \\
\makecell[l]{PSMNet \cite{chang2018pyramid}} 			& 16.3\% & 15.1\% & 34.2\% & 23.8\%  \\
\makecell[l]{GANet \cite{zhang2019ga}} 				& 11.7\% & 10.1\% & 20.3\% & 14.1\% \\
\makecell[l]{MS-PSMNet \cite{cai2020matching}} 			& 7.8\%  & 14.0\%  & 19.8\% & 16.8\%  \\
\makecell[l]{MS-GCNet \cite{cai2020matching}} 			& 6.2\% & 5.5\% & 18.5\% & \underline{8.8\%} \\
\makecell[l]{DSMNet \cite{zhang2020domain}}		        & 6.5\% &  6.2\% & 13.8\% & \textbf{6.2\%} \\
\makecell[l]{Graft-PSMNet} 						& \textbf{5.3\%} & \underline{5.0\%} & \underline{10.9\%} & 10.7\% \\
\makecell[l]{Graft-GANet} 							& \underline{5.4\%} & \textbf{4.6\%} & \textbf{8.9\%} & \textbf{6.2\%} \\
\midrule
\makecell[l]{CFNet* \cite{shen2021cfnet}} 				& 6.0\% & 5.1\% & 15.4\% & 5.3\% \\
\makecell[l]{RAFT-Stereo* \cite{lipson2021raft}} 			& 5.7\% & - & 12.6\% & \textbf{3.3\%} \\
\makecell[l]{SGM+NDR \cite{aleotti2021neural}}			& 5.5\% & 6.0\% & 12.4\% & \underline{4.8\%} \\
\makecell[l]{Graft-PSMNet*} 						& \textbf{4.8\%} & \underline{4.3\%} & \textbf{9.7\%} & 7.7\% \\
\makecell[l]{Graft-GANet*} 							& \underline{4.9\%} & \textbf{4.2\%} & \underline{9.8\%} & 6.2\% \\
\bottomrule
\end{tabular}
\caption{Comparison of robust and domain generalized stereo matching methods, ours are listed at the bottom of the two subtables. * means the color jitter data augmentation strategy is leveraged during training. \textbf{KT-15}: KITTI 2015, \textbf{KT-12}: KITTI 2012, \textbf{MB}: Middlebury, \textbf{ET}: ETH3D. The best result is shown in \textbf{bold} and the second result is \underline{underlined}.}
\label{comparison}
\end{table}

\section{Limitation \& Future Work}

There are two main limitations of our work: 1) As discussed in Section \ref{section4.6}, the grafted feature is not perfectly domain-invariant. 2) The annotations for image classification are implicitly used through loading the parameters of the model trained on ImageNet, meaning that additional labeled data (but not limited to stereo matching) are needed. 

Aiming at these limitations, we intend to deeply combine self-supervised representation learning with our GraftNet in the future. By this means, only images are required and huge amounts of data from the Internet can be leveraged to improve the robustness of the learned feature. 

\section{Conclusion}

This paper attempts to achieve domain generalized stereo matching from the perspective of \emph{data}, where the key is a broad-spectrum and task-oriented feature. The former property comes from the various styles of images seen during training, and the latter property is realized by recovering task-related information from the broad-spectrum feature. By constructing a generalized cost space with cosine similarity, the feature is combined with an ordinary cost aggregation module. Experimental results on several datasets show that our Graft-PSMNet and Graft-GANet are superior to other robust and domain generalized algorithms. We hope our method can inspire subsequent studies, including multi-task learning and domain generalized approaches.

\section*{Societal Impact}

The fundamental purpose of our work is to improve the quality of the depth obtained by stereo systems, which plays an essential role in many industries such as robot navigation and autonomous driving. On the one hand, the developments of these industries bring huge convenience to human activities. On the other hand, there are still lots of questions to be answered about security, emotion, etc. To handle these potential problems, we must not only carefully evaluate the security of the AI (Artificial Intelligence) systems, but also establish and complete some related laws.

{\small
\bibliographystyle{ieee_fullname}
\bibliography{egbib}

\begin{thebibliography}{10}\itemsep=-1pt

\bibitem{aleotti2021neural}
Filippo Aleotti, Fabio Tosi, Pierluigi~Zama Ramirez, Matteo Poggi, Samuele
  Salti, Stefano Mattoccia, and Luigi Di~Stefano.
\newblock Neural disparity refinement for arbitrary resolution stereo.
\newblock {\em arXiv preprint arXiv:2110.15367}, 2021.

\bibitem{cai2020matching}
Changjiang Cai, Matteo Poggi, Stefano Mattoccia, and Philippos Mordohai.
\newblock Matching-space stereo networks for cross-domain generalization.
\newblock In {\em 2020 International Conference on 3D Vision (3DV)}, pages
  364--373. IEEE, 2020.

\bibitem{chang2018pyramid}
Jia-Ren Chang and Yong-Sheng Chen.
\newblock Pyramid stereo matching network.
\newblock In {\em Proceedings of the IEEE Conference on Computer Vision and
  Pattern Recognition}, pages 5410--5418, 2018.

\bibitem{chen2017deeplab}
Liang-Chieh Chen, George Papandreou, Iasonas Kokkinos, Kevin Murphy, and Alan~L
  Yuille.
\newblock Deeplab: Semantic image segmentation with deep convolutional nets,
  atrous convolution, and fully connected crfs.
\newblock {\em IEEE transactions on pattern analysis and machine intelligence},
  40(4):834--848, 2017.

\bibitem{chen2015deep}
Zhuoyuan Chen, Xun Sun, Liang Wang, Yinan Yu, and Chang Huang.
\newblock A deep visual correspondence embedding model for stereo matching
  costs.
\newblock In {\em Proceedings of the IEEE International Conference on Computer
  Vision}, pages 972--980, 2015.

\bibitem{cheng2020hierarchical}
Xuelian Cheng, Yiran Zhong, Mehrtash Harandi, Yuchao Dai, Xiaojun Chang, Tom
  Drummond, Hongdong Li, and Zongyuan Ge.
\newblock Hierarchical neural architecture search for deep stereo matching.
\newblock {\em arXiv preprint arXiv:2010.13501}, 2020.

\bibitem{deng2009imagenet}
Jia Deng, Wei Dong, Richard Socher, Li-Jia Li, Kai Li, and Li Fei-Fei.
\newblock Imagenet: A large-scale hierarchical image database.
\newblock In {\em 2009 IEEE conference on computer vision and pattern
  recognition}, pages 248--255. Ieee, 2009.

\bibitem{geiger2012we}
Andreas Geiger, Philip Lenz, and Raquel Urtasun.
\newblock Are we ready for autonomous driving? the kitti vision benchmark
  suite.
\newblock In {\em 2012 IEEE Conference on Computer Vision and Pattern
  Recognition}, pages 3354--3361. IEEE, 2012.

\bibitem{godard2019digging}
Cl{\'e}ment Godard, Oisin Mac~Aodha, Michael Firman, and Gabriel~J Brostow.
\newblock Digging into self-supervised monocular depth estimation.
\newblock In {\em Proceedings of the IEEE/CVF International Conference on
  Computer Vision}, pages 3828--3838, 2019.

\bibitem{guo2019group}
Xiaoyang Guo, Kai Yang, Wukui Yang, Xiaogang Wang, and Hongsheng Li.
\newblock Group-wise correlation stereo network.
\newblock In {\em Proceedings of the IEEE/CVF Conference on Computer Vision and
  Pattern Recognition}, pages 3273--3282, 2019.

\bibitem{he2016deep}
Kaiming He, Xiangyu Zhang, Shaoqing Ren, and Jian Sun.
\newblock Deep residual learning for image recognition.
\newblock In {\em Proceedings of the IEEE conference on computer vision and
  pattern recognition}, pages 770--778, 2016.

\bibitem{hirschmuller2005accurate}
Heiko Hirschmuller.
\newblock Accurate and efficient stereo processing by semi-global matching and
  mutual information.
\newblock In {\em 2005 IEEE Computer Society Conference on Computer Vision and
  Pattern Recognition (CVPR'05)}, volume~2, pages 807--814. IEEE, 2005.

\bibitem{hirschmuller2007evaluation}
Heiko Hirschmuller and Daniel Scharstein.
\newblock Evaluation of cost functions for stereo matching.
\newblock In {\em 2007 IEEE Conference on Computer Vision and Pattern
  Recognition}, pages 1--8. IEEE, 2007.

\bibitem{kendall2017end}
Alex Kendall, Hayk Martirosyan, Saumitro Dasgupta, Peter Henry, Ryan Kennedy,
  Abraham Bachrach, and Adam Bry.
\newblock End-to-end learning of geometry and context for deep stereo
  regression.
\newblock In {\em Proceedings of the IEEE International Conference on Computer
  Vision}, pages 66--75, 2017.

\bibitem{li2020knowledge}
Wei-Hong Li and Hakan Bilen.
\newblock Knowledge distillation for multi-task learning.
\newblock In {\em European Conference on Computer Vision}, pages 163--176.
  Springer, 2020.

\bibitem{li2021revisiting}
Zhaoshuo Li, Xingtong Liu, Nathan Drenkow, Andy Ding, Francis~X Creighton,
  Russell~H Taylor, and Mathias Unberath.
\newblock Revisiting stereo depth estimation from a sequence-to-sequence
  perspective with transformers.
\newblock In {\em Proceedings of the IEEE/CVF International Conference on
  Computer Vision}, pages 6197--6206, 2021.

\bibitem{lipson2021raft}
Lahav Lipson, Zachary Teed, and Jia Deng.
\newblock Raft-stereo: Multilevel recurrent field transforms for stereo
  matching.
\newblock {\em arXiv preprint arXiv:2109.07547}, 2021.

\bibitem{liu2020stereogan}
Rui Liu, Chengxi Yang, Wenxiu Sun, Xiaogang Wang, and Hongsheng Li.
\newblock Stereogan: Bridging synthetic-to-real domain gap by joint
  optimization of domain translation and stereo matching.
\newblock In {\em Proceedings of the IEEE/CVF conference on computer vision and
  pattern recognition}, pages 12757--12766, 2020.

\bibitem{luo2016efficient}
Wenjie Luo, Alexander~G Schwing, and Raquel Urtasun.
\newblock Efficient deep learning for stereo matching.
\newblock In {\em Proceedings of the IEEE conference on computer vision and
  pattern recognition}, pages 5695--5703, 2016.

\bibitem{mayer2016large}
Nikolaus Mayer, Eddy Ilg, Philip Hausser, Philipp Fischer, Daniel Cremers,
  Alexey Dosovitskiy, and Thomas Brox.
\newblock A large dataset to train convolutional networks for disparity,
  optical flow, and scene flow estimation.
\newblock In {\em Proceedings of the IEEE conference on computer vision and
  pattern recognition}, pages 4040--4048, 2016.

\bibitem{menze2015object}
Moritz Menze and Andreas Geiger.
\newblock Object scene flow for autonomous vehicles.
\newblock In {\em Proceedings of the IEEE conference on computer vision and
  pattern recognition}, pages 3061--3070, 2015.

\bibitem{paszke2017automatic}
Adam Paszke, Sam Gross, Soumith Chintala, Gregory Chanan, Edward Yang, Zachary
  DeVito, Zeming Lin, Alban Desmaison, Luca Antiga, and Adam Lerer.
\newblock Automatic differentiation in pytorch.
\newblock 2017.

\bibitem{poggi2019guided}
Matteo Poggi, Davide Pallotti, Fabio Tosi, and Stefano Mattoccia.
\newblock Guided stereo matching.
\newblock In {\em Proceedings of the IEEE/CVF Conference on Computer Vision and
  Pattern Recognition}, pages 979--988, 2019.

\bibitem{ramirez2019learning}
Pierluigi~Zama Ramirez, Alessio Tonioni, Samuele Salti, and Luigi~Di Stefano.
\newblock Learning across tasks and domains.
\newblock In {\em Proceedings of the IEEE/CVF International Conference on
  Computer Vision}, pages 8110--8119, 2019.

\bibitem{ronneberger2015u}
Olaf Ronneberger, Philipp Fischer, and Thomas Brox.
\newblock U-net: Convolutional networks for biomedical image segmentation.
\newblock In {\em International Conference on Medical image computing and
  computer-assisted intervention}, pages 234--241. Springer, 2015.

\bibitem{scharstein2014high}
Daniel Scharstein, Heiko Hirschm{\"u}ller, York Kitajima, Greg Krathwohl, Nera
  Ne{\v{s}}i{\'c}, Xi Wang, and Porter Westling.
\newblock High-resolution stereo datasets with subpixel-accurate ground truth.
\newblock In {\em German conference on pattern recognition}, pages 31--42.
  Springer, 2014.

\bibitem{scharstein2002taxonomy}
Daniel Scharstein and Richard Szeliski.
\newblock A taxonomy and evaluation of dense two-frame stereo correspondence
  algorithms.
\newblock {\em International journal of computer vision}, 47(1):7--42, 2002.

\bibitem{schoeps2017cvpr}
Thomas Sch\"ops, Johannes~L. Sch\"onberger, Silvano Galliani, Torsten Sattler,
  Konrad Schindler, Marc Pollefeys, and Andreas Geiger.
\newblock A multi-view stereo benchmark with high-resolution images and
  multi-camera videos.
\newblock In {\em Conference on Computer Vision and Pattern Recognition
  (CVPR)}, 2017.

\bibitem{shen2021cfnet}
Zhelun Shen, Yuchao Dai, and Zhibo Rao.
\newblock Cfnet: Cascade and fused cost volume for robust stereo matching.
\newblock In {\em Proceedings of the IEEE/CVF Conference on Computer Vision and
  Pattern Recognition}, pages 13906--13915, 2021.

\bibitem{simonyan2014very}
Karen Simonyan and Andrew Zisserman.
\newblock Very deep convolutional networks for large-scale image recognition.
\newblock {\em arXiv preprint arXiv:1409.1556}, 2014.

\bibitem{song2021adastereo}
Xiao Song, Guorun Yang, Xinge Zhu, Hui Zhou, Zhe Wang, and Jianping Shi.
\newblock Adastereo: a simple and efficient approach for adaptive stereo
  matching.
\newblock In {\em Proceedings of the IEEE/CVF Conference on Computer Vision and
  Pattern Recognition}, pages 10328--10337, 2021.

\bibitem{song2020edgestereo}
Xiao Song, Xu Zhao, Liangji Fang, Hanwen Hu, and Yizhou Yu.
\newblock Edgestereo: An effective multi-task learning network for stereo
  matching and edge detection.
\newblock {\em International Journal of Computer Vision}, 128(4):910--930,
  2020.

\bibitem{teed2020raft}
Zachary Teed and Jia Deng.
\newblock Raft: Recurrent all-pairs field transforms for optical flow.
\newblock In {\em European conference on computer vision}, pages 402--419.
  Springer, 2020.

\bibitem{tonioni2019learning}
Alessio Tonioni, Oscar Rahnama, Thomas Joy, Luigi~Di Stefano, Thalaiyasingam
  Ajanthan, and Philip~HS Torr.
\newblock Learning to adapt for stereo.
\newblock In {\em Proceedings of the IEEE/CVF Conference on Computer Vision and
  Pattern Recognition}, pages 9661--9670, 2019.

\bibitem{tonioni2019real}
Alessio Tonioni, Fabio Tosi, Matteo Poggi, Stefano Mattoccia, and Luigi~Di
  Stefano.
\newblock Real-time self-adaptive deep stereo.
\newblock In {\em Proceedings of the IEEE/CVF Conference on Computer Vision and
  Pattern Recognition}, pages 195--204, 2019.

\bibitem{tulyakov2018practical}
Stepan Tulyakov, Anton Ivanov, and Fran{\c{c}}ois Fleuret.
\newblock Practical deep stereo (pds): Toward applications-friendly deep stereo
  matching.
\newblock {\em Advances in Neural Information Processing Systems},
  31:5871--5881, 2018.

\bibitem{wang2021dense}
Xinlong Wang, Rufeng Zhang, Chunhua Shen, Tao Kong, and Lei Li.
\newblock Dense contrastive learning for self-supervised visual pre-training.
\newblock In {\em Proceedings of the IEEE/CVF Conference on Computer Vision and
  Pattern Recognition}, pages 3024--3033, 2021.

\bibitem{wang2020generalizing}
Yaqing Wang, Quanming Yao, James~T Kwok, and Lionel~M Ni.
\newblock Generalizing from a few examples: A survey on few-shot learning.
\newblock {\em ACM Computing Surveys (CSUR)}, 53(3):1--34, 2020.

\bibitem{watson2020learning}
Jamie Watson, Oisin Mac~Aodha, Daniyar Turmukhambetov, Gabriel~J Brostow, and
  Michael Firman.
\newblock Learning stereo from single images.
\newblock In {\em European Conference on Computer Vision}, pages 722--740.
  Springer, 2020.

\bibitem{xu2020aanet}
Haofei Xu and Juyong Zhang.
\newblock Aanet: Adaptive aggregation network for efficient stereo matching.
\newblock In {\em Proceedings of the IEEE/CVF Conference on Computer Vision and
  Pattern Recognition}, pages 1959--1968, 2020.

\bibitem{yang2018segstereo}
Guorun Yang, Hengshuang Zhao, Jianping Shi, Zhidong Deng, and Jiaya Jia.
\newblock Segstereo: Exploiting semantic information for disparity estimation.
\newblock In {\em Proceedings of the European Conference on Computer Vision
  (ECCV)}, pages 636--651, 2018.

\bibitem{yoon2006adaptive}
Kuk-Jin Yoon and In~So Kweon.
\newblock Adaptive support-weight approach for correspondence search.
\newblock {\em IEEE transactions on pattern analysis and machine intelligence},
  28(4):650--656, 2006.

\bibitem{zbontar2015computing}
Jure Zbontar and Yann LeCun.
\newblock Computing the stereo matching cost with a convolutional neural
  network.
\newblock In {\em Proceedings of the IEEE conference on computer vision and
  pattern recognition}, pages 1592--1599, 2015.

\bibitem{zhang2019ga}
Feihu Zhang, Victor Prisacariu, Ruigang Yang, and Philip~HS Torr.
\newblock Ga-net: Guided aggregation net for end-to-end stereo matching.
\newblock In {\em Proceedings of the IEEE/CVF Conference on Computer Vision and
  Pattern Recognition}, pages 185--194, 2019.

\bibitem{zhang2020domain}
Feihu Zhang, Xiaojuan Qi, Ruigang Yang, Victor Prisacariu, Benjamin Wah, and
  Philip Torr.
\newblock Domain-invariant stereo matching networks.
\newblock In {\em European Conference on Computer Vision}, pages 420--439.
  Springer, 2020.

\end{thebibliography}
}

\end{document}